\pdfoutput=1

\documentclass[11pt]{article}

\usepackage{EMNLP2023}

\usepackage{times}
\usepackage{latexsym}

\usepackage[T1]{fontenc}

\usepackage[utf8]{inputenc}

\usepackage{microtype}

\usepackage{inconsolata}

\usepackage{amsmath}
\usepackage{graphicx}
\usepackage{booktabs}
\usepackage{amssymb}
\usepackage{transparent}
\usepackage{multirow}
\usepackage{tikz}
\usepackage[capitalize]{cleveref}
\usepackage{pifont}
\usepackage[super]{nth}
\usepackage[shortlabels]{enumitem}
\usepackage{quoting}
\usepackage{subcaption}
\usepackage{adjustbox}

\usepackage{array}
\usepackage{booktabs}
\newcommand{\minitab}[2][l]{\begin{tabular}{#1}#2\end{tabular}}

\crefformat{section}{\S#2#1#3} 
\crefformat{subsection}{\S#2#1#3}
\crefformat{subsubsection}{\S#2#1#3}

\mathchardef\mhyphen="2D 

\title{A Comparative Analysis of Task-Agnostic Distillation Methods \\for Compressing Transformer Language Models}

\author{Takuma Udagawa, Aashka Trivedi, Michele Merler,  Bishwaranjan Bhattacharjee \\
IBM Research AI \\
\texttt{\{takuma.udagawa@, aashka.trivedi@, mimerler@us., bhatta@us.\}ibm.com}}

\begin{document}
\maketitle
\begin{abstract}
Large language models have become a vital component in modern NLP, achieving state of the art performance in a variety of tasks.
However, they are often inefficient for real-world deployment due to their expensive inference costs.
Knowledge distillation is a promising technique to improve their efficiency while retaining most of their effectiveness.
In this paper, we reproduce, compare and analyze several representative methods for task-agnostic (general-purpose) distillation of Transformer language models.
Our target of study includes Output Distribution (OD) transfer, Hidden State (HS) transfer with various layer mapping strategies, and Multi-Head Attention (MHA) transfer based on MiniLMv2.
Through our extensive experiments, we study the effectiveness of each method for various student architectures in both monolingual (English) and multilingual settings.
Overall, we show that MHA transfer based on MiniLMv2 is generally the best option for distillation and explain the potential reasons behind its success.
Moreover, we show that HS transfer remains as a competitive baseline, especially under a sophisticated layer mapping strategy, while OD transfer consistently lags behind other approaches.
Findings from this study helped us deploy efficient yet effective student models for latency-critical applications.
\end{abstract}

\section{Introduction}
\label{sec:introduction}

\begin{figure*}[t!]
    \centering
    \begin{subfigure}[b]{0.22\textwidth}
        \centering
        \includegraphics[height=0.7in]{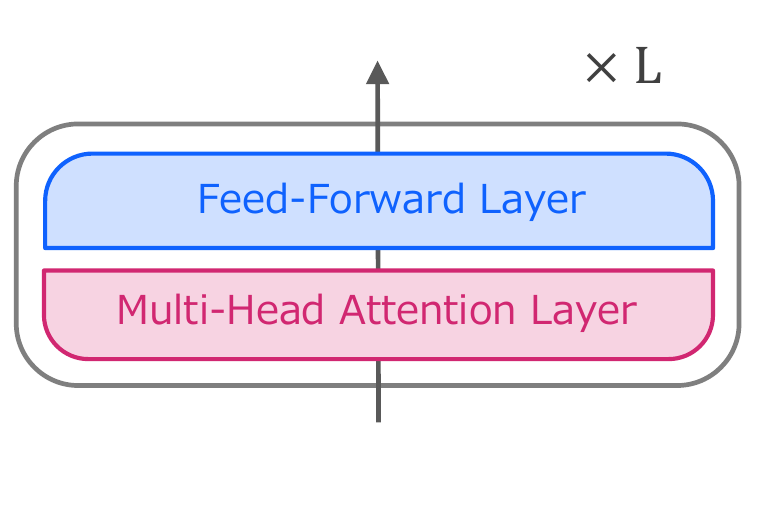}
        \vspace{0.15in}
        \caption{Transformer LM}
    \end{subfigure}
    \begin{subfigure}[b]{0.25\textwidth}
        \centering
        \includegraphics[height=1.1in]{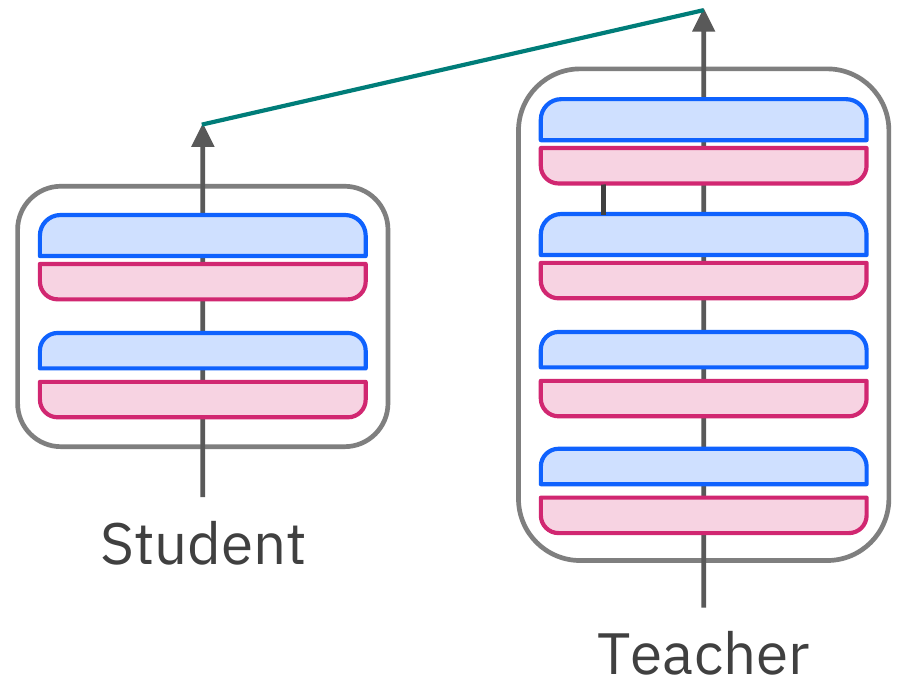}
        \caption{OD Transfer}
    \end{subfigure}
    \begin{subfigure}[b]{0.25\textwidth}
        \centering
        \includegraphics[height=1.1in]{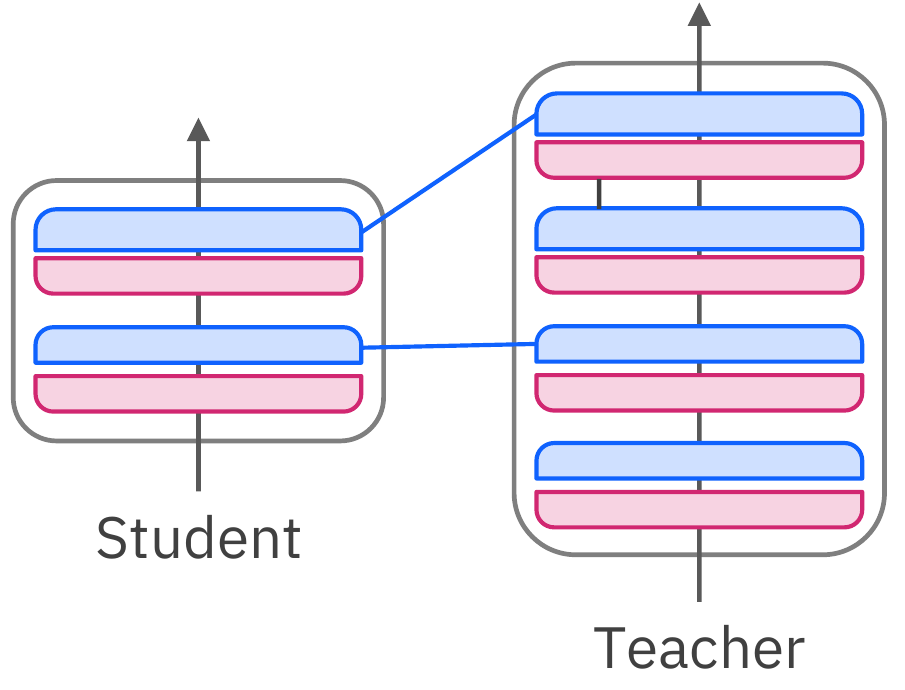}
        \caption{HS Transfer}
    \end{subfigure}
    \begin{subfigure}[b]{0.25\textwidth}
        \centering
        \includegraphics[height=1.1in]{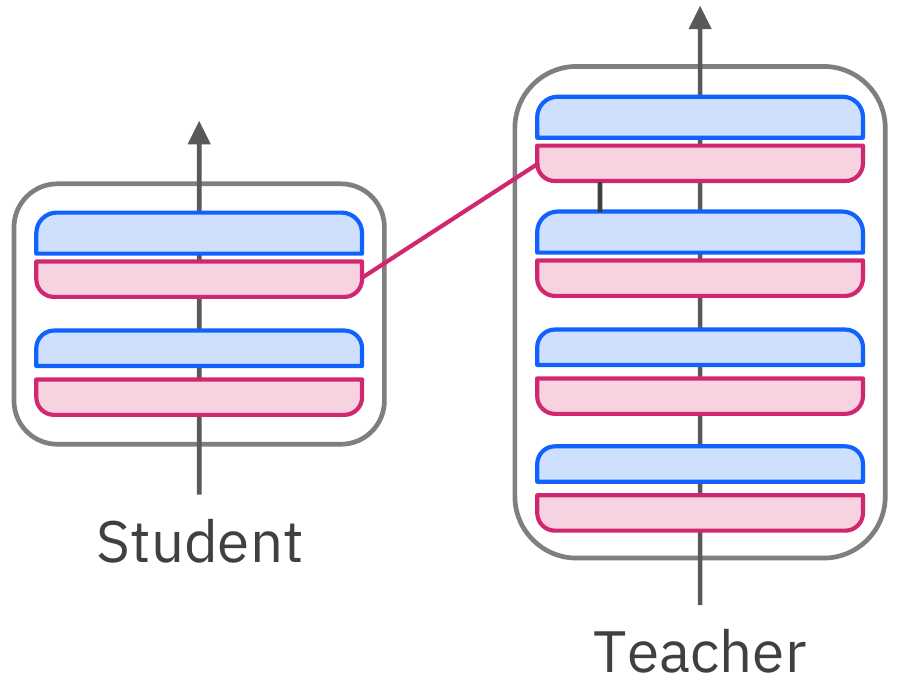}
        \caption{MHA Transfer}
    \end{subfigure}
    \caption{A high-level illustration of (a) the Transformer architecture and (b-d) representative distillation methods. (b-d) denote Output Distribution (OD), Hidden State (HS), and Multi-Head Attention (MHA) transfer, respectively. Lines between the student and teacher depict which level of information is transferred in each method.}
    \label{fig:summary}
\end{figure*}

Large language models have become a crucial component in modern NLP.
They have achieved exceptional performance on various downstream tasks \citep{devlin2019bert,liu2019roberta,lewis-etal-2020-bart} and their capability shows consistent improvement with more compute, data, and model parameters \citep{kaplan2020scaling,NEURIPS2020_1457c0d6,touvron2023llama}.
On the downside, it is becoming increasingly difficult to deploy such models in real-world environments due to their \textit{inefficiency}, i.e. high computation, memory, latency and storage costs \citep{xu2023survey}.

Knowledge distillation \citep{Hinton2015DistillingTK} is a promising technique to overcome this challenge by transferring the knowledge of the original model (teacher) to a smaller, more efficient model (student).
This can be conducted in either \textit{task-specific} \citep{turc2019well,jiao2020tinybert} or \textit{task-agnostic} manner \citep{sanh2019distilbert,wang2020minilm}.
The latter only requires distilling a single general-purpose student which can be directly finetuned on any downstream task. Due to its high convenience, we focus on this latter approach in this study.

In recent years, there have been various methods proposed for task-agnostic distillation of Transformer language models. 
The aim of this paper is to reproduce, compare and analyze the most representative methods in this area.
We generally focus on the \textit{architecture-agnostic} distillation which imposes no or minimal restriction on the student architecture\footnote{By \textit{architecture-agnostic}, we mean that the student and teacher can have different architectural parameters (e.g. number of layers, attention heads, hidden state size, etc).}:
the representative methods include Output Distribution (OD) transfer \citep{Hinton2015DistillingTK}, Hidden State (HS) transfer based on linear mapping \citep{jiao2020tinybert,mukherjee2021xtremedistiltransformers} and Multi-Head Attention (MHA) transfer based on MiniLMv2 \citep{wang-etal-2021-minilmv2}.

For HS transfer, the layer mapping strategy between teacher and student layers plays a significant role in overall performance, however, the optimal strategy remains unknown or controversial  \citep{sun-etal-2019-patient,wu-etal-2020-skip,ko2023revisiting}. Therefore, we explore a diverse range of strategies to empirically evaluate each technique.

For MHA transfer, the MiniLMv2 approach has been shown to achieve state-of-the-art performance, however, there is relatively little understanding behind its success.
Therefore, we develop a novel variant named DirectMiniLM which is useful for understanding the effectiveness behind MiniLMv2 both theoretically and empirically.

In contrast to most previous studies, all methods are reproduced on a single unified codebase for fair and consistent comparison.
We also conduct distillation on 4 different student architectures, reducing the model size in various dimensions to fit various parameter and latency budgets. 
Finally, all experiments are conducted on both monolingual and multilingual settings, distilled from open-source BERT \cite{devlin2019bert} and in-house XLM-RoBERTa \cite{conneau2020unsupervised}, respectively.

Through our extensive experiments, we critically analyze the effectiveness of each distillation method and provide practical advice for both researchers and practitioners working in this area.

In summary, our key findings are:

\begin{itemize}[topsep=0pt, itemsep=0pt, leftmargin=.2in, parsep=0pt]
    \item MHA transfer is generally the best option for various student architectures and language settings. By comparison with DirectMiniLM, we provide novel insights underlying its success.
    \item While the effectiveness of HS transfer depends on the layer mapping strategy, it remains as a competitive baseline. More sophisticated layer mapping strategy can provide a boost in performance, esp. in the multilingual setting.
    \item Methods relying on OD transfer consistently lag behind other methods. This shows that classical OD distillation can be less effective when distilling complex language models on a general-purpose objective.
\end{itemize}

\section{Transformer Language Models}
\label{sec:transformer_lms}

First, we briefly review the standard architecture of Transformer language models \citep{vaswani2017attention,devlin2019bert}.
A Transformer consists of a stack of $L$ Transformer layers, where each layer comprises two sub-layers: a Multi-Head Attention (MHA) layer followed by a fully connected Feed-Forward (FF) layer (Figure \ref{fig:summary}, (a)).

Formally, let $x$ denote the input sequence, $d_h$ the hidden state size, and $\mathbf{H}_{i} \in \mathbb{R}^{|x| \times d_h}$ the hidden state of the $i^{\mathrm{th}}$ Transformer layer ($\mathbf{H}_0$ denotes the input sequence embeddings).
Given $\mathbf{H}_i$, the MHA layer first computes the query, key, and value mappings $\mathbf{Q}_{i,a}$, $\mathbf{K}_{i,a}$, $\mathbf{V}_{i,a}$ for each attention head $a \in [1,A_h]$, which are combined to obtain the attention head output $\mathbf{O}_{i,a}$:
\begin{gather}
\mathbf{Q}_{i,a} = \mathbf{H}_{i} \mathbf{W}_{Q,i,a} \\
\mathbf{K}_{i,a} = \mathbf{H}_{i} \mathbf{W}_{K,i,a} \\ 
\mathbf{V}_{i,a} = \mathbf{H}_{i} \mathbf{W}_{V,i,a} \\
\mathbf{O}_{i,a} = \mathrm{softmax}(\frac{\mathbf{Q}_{i,a}\, \mathbf{K}_{i,a}^{\!\mathsf{T}}}{ \sqrt{d_k}})\mathbf{V}_{i,a}
\end{gather}
\noindent
Here, $d_k$ denotes the attention head size (typically set to $\frac{d_h}{A_h}$) and $\mathbf{W}_{Q,i,a}, \mathbf{W}_{K,i,a}, \mathbf{W}_{V,i,a} \in \mathbb{R}^{d_h \times d_k}$ are the learnt weight matrices.
The output of the MHA layer is the concatenation of $\mathbf{O}_{i,a}$, namely $\mathrm{MHA}(\mathbf{H}_{i}) = \bigoplus_{a=1}^{A_h} \mathbf{O}_{i,a}$.

Next, the MHA layer output is followed by a position-wise FF layer with an intermediate size of $d_f$ and a non-linear activation (we use GELU \citep{hendrycks2016gaussian} in all models). The hidden state of the next Transformer layer is computed as $\mathbf{H}_{i+1} = \mathrm{FF}(\mathrm{MHA}(\mathbf{H}_{i}))$.\footnote{Both MHA and FF sub-layers have a residual connection \citep{he2016deep} and are followed by layer normalization \citep{ba2016layer}, which are omitted for brevity.}

Finally, to predict the output distribution over the entire vocabulary $V$, a linear layer $\mathbf{W}_{O} \in  \mathbb{R}^{d_h \times |V|}$ is applied on top of the last hidden state to compute the logits $\mathbf{z} = \mathbf{H}_{L} \mathbf{W}_{O} \in \mathbb{R}^{|x| \times |V|}$.
The output distribution can be obtained by applying the softmax function over $\mathbf{z}$, denoted as $\mathrm{softmax} (\mathbf{z})$.

Throughout this paper, we assume that both the student and teacher are Transformer language models with $L^S$ and $L^T$ layers, respectively.

\section{Distillation Methods}
\label{sec:distillation_methods}

Next, we introduce the representative task-agnostic distillation methods illustrated in Figure \ref{fig:summary}, (b-d).
For Multi-Head Attention (MHA) transfer, we consider two approaches: MiniLMv2 and its novel variant DirectMiniLM.
For a survey of advanced methods and topics we could not cover in this study, please refer to Appendix \ref{sec:related_work}.

\paragraph{Output Distribution (OD) Transfer}
The output distribution of the teacher contains useful information on the relative probabilities of plausible (even if incorrect) predictions \citep{Hinton2015DistillingTK}.
In OD transfer, the student is trained to replicate the teacher's output distribution.
This is achieved by optimizing the following loss function, where $\mathbf{z}^S, \mathbf{z}^T$ denote the student/teacher logits, $\mathrm{CE}(.)$ the cross entropy loss and $\mathcal{T}$ the output temperature:
\begin{equation}
\mathcal{L}_{\mathrm{OD}} = \mathcal{T}^2 \cdot \mathrm{CE} \Bigl(\mathrm{softmax} \bigl( \frac{\mathbf{z}^T}{\mathcal{T}} \bigl), \mathrm{softmax} \bigl( \frac{\mathbf{z}^S}{\mathcal{T}} \bigl) \Bigl) 
\label{eq:od_transfer}
\end{equation}

\noindent

\paragraph{Hidden State (HS) Transfer} Transformer language models progressively learn useful and generalizable features layer by layer.
In HS transfer, the student is trained to predict such useful features represented in the teacher's hidden states.

Formally, each student layer is mapped to a set of teacher layers to be predicted.
Let $\phi(i)$ denote the set mapped from the $i^{\mathrm{th}}$ student layer, where $\emptyset \subseteq \phi(i) \subseteq [1,L^T]$.
For each $j \in \phi(i)$, the hidden state of the $i^{\mathrm{th}}$ student layer $\mathbf{H}^S_{i} \in \mathbb{R}^{|x| \times {d}^S_{h}}$ is linearly transformed to predict the hidden state of the $j^{\mathrm{th}}$ teacher layer $\mathbf{H}^T_{j} \in \mathbb{R}^{|x| \times {d}^T_{h}}$.\footnote{Note that ${d}^S_{h}$ and ${d}^T_{h}$ are the student and teacher hidden state sizes which can take different values.}
This is represented by the following loss function, where $\mathbf{W}_i^j \in \mathbb{R}^{{d}^S_{h} \times {d}^T_{h}}$ denotes the linear transformation weight and $\mathrm{MSE}(.)$ the mean squared error loss:
\begin{equation}
\mathcal{L}_{\mathrm{HS}} = \sum_{i=1}^{L^S} \sum_{j \in \phi(i)} \mathrm{MSE} \Bigl( \mathbf{H}^S_i \mathbf{W}_i^j, \mathbf{H}^T_{j} \Bigl)
\label{eq:hs_transfer}
\end{equation}

One open problem in this approach is the choice of layer mapping strategy $\phi$.
We conduct extensive experiments to compare a diverse range of strategies, which will be discussed in \cref{sec:experiments}.

\paragraph{MiniLMv2} The MHA layer is a key component in Transformer language models which controls the long-range dependencies and interactions within input texts.
MiniLMv2 \citep{wang-etal-2021-minilmv2} is an effective method to deeply transfer this module while allowing different number of attention heads ${A}^S_{h}$ and ${A}^T_{h}$ for the student and teacher.
Their main idea is to distil the attention \textit{relation} matrices (Q-Q, K-K and V-V) obtained by first concatenating the query (Q), key (K), and value (V) mappings from all attention heads and re-splitting them into the same number of attention \textit{relation} heads $A_r$.

Formally, let $\mathbf{A}_{Q, i, a}^S, \mathbf{A}_{K, i, a}^S, \mathbf{A}_{V, i, a}^S \in \mathbb{R}^{|x| \times {d}^S_{r}}$ denote the concatenated and re-split queries, keys, and values for the $i^{\mathrm{th}}$ student layer, where $a \in [1,A_r]$ and ${d}^S_{r} = \frac{{d}^S_{h}}{A_r}$. For instance, $\bigoplus_{a=1}^{{A}^S_{h}} \mathbf{Q}_{i, a}^S = \bigoplus_{a=1}^{A_r} \mathbf{A}_{Q, i, a}^S$, i.e. original queries from ${A}^S_{h}$ attention heads are simply concatenated and then re-split into $A_r$ matrices.
We use the same notation for the $j^{\mathrm{th}}$ teacher layer, $\mathbf{A}_{Q, j, a}^T, \mathbf{A}_{K, j, a}^T, \mathbf{A}_{V, j, a}^T \in \mathbb{R}^{|x| \times {d}^T_{r}}$, where ${d}^T_{r} = \frac{{d}^T_{h}}{A_r}$.
Then, the loss function of MiniLMv2 can be defined as follows:\begin{equation}
\mathcal{L}_{\mathrm{MHA}} = \!\! \sum_{\alpha \in \{ Q,K,V \}} \sum_{a = 1}^{A_r} \mathrm{CE} \Bigl( \mathbf{R}^T_{\alpha, j, a},  \mathbf{R}^S_{\alpha, i, a} \Bigl)
\label{eq:mha_transfer}
\end{equation}
\begin{equation}
\mathbf{R}_{\alpha, j, a}^T = \mathrm{softmax} \Bigl( \frac{\mathbf{A}_{\alpha, j, a}^T \mathbf{A}_{\alpha, j, a}^{T\mathsf{\,T}}}{\sqrt{{d}^T_{r}}} \Bigl)
\label{eq:rt_transfer}
\end{equation}
\begin{equation}
\mathbf{R}_{\alpha, i, a}^S = \mathrm{softmax} \Bigl( \frac{\mathbf{A}_{\alpha, i, a}^S \mathbf{A}_{\alpha, i, a}^{S\mathsf{\,T}}}{\sqrt{{d}^S_{r}}} \Bigl)
\label{eq:rs_transfer}
\end{equation}

\noindent
Here, $\mathbf{R}_{\alpha, j, a}^T, \mathbf{R}_{\alpha, i, a}^S \in \mathbb{R}^{|x| \times |x|}$ denote the attention \textit{relation} matrices which are computed based on the matrix products of $\mathbf{A}^T_{\alpha, i, a}, \mathbf{A}^S_{\alpha, i, a}$ in eq. (\ref{eq:rt_transfer}), (\ref{eq:rs_transfer}), respectively.
Intuitively, this aims to transfer the teacher's queries (Q), keys (K) and values (V) in a somewhat indirect way through their matrix products (Q-Q, K-K and V-V).

However, there is minimal justification for why this method works effectively.
It is also difficult to directly compare the method against HS transfer since the losses are computed differently.
To better understand MiniLMv2, we propose its novel variant named DirectMiniLM for our analysis.

\paragraph{DirectMiniLM}

In DirectMiniLM, we aim to transfer the teacher's Q/K/V mappings more directly through the linear transformation of the student's ones, just as we did in HS transfer.
Specifically, we use the following loss function with the linear transformation $\mathbf{W}_{\alpha, a} \in \mathbb{R}^{d^S_r \times d^T_r}$:\begin{equation}
\mathcal{L}_{\mathrm{MHA}}^{\mathrm{Direct}} =\!\! \sum_{\substack{\alpha \, \in \\ \{ Q,K,V \}}} \sum_{a = 1}^{A_r} \mathrm{MSE} \Bigl( \mathbf{A}_{\alpha, i, a}^S \mathbf{W}_{\alpha, a}, \mathbf{A}_{\alpha, j, a}^T \Bigl)
\label{eq:direct_minilm}
\end{equation}

DirectMiniLM is important in two aspects.
First, this approach is directly comparable to HS transfer based on eq. (\ref{eq:hs_transfer}) with the only difference in which information you transfer: the hidden states $\mathbf{H}^T_i \rightarrow \mathbf{H}^S_j$ or the Q/K/V mappings  $\mathbf{A}_{\alpha, i, a}^T \rightarrow \mathbf{A}_{\alpha, j, a}^S$.
From this comparison, we can quantify the precise advantage of transferring each knowledge in an apples-to-apples manner. 

Second, DirectMiniLM is also closely relevant to MiniLMv2: if we constrain $\mathbf{W}_{\alpha, a}$ to be orthogonal (i.e. $\mathbf{W}_{\alpha, a} \mathbf{W}_{\alpha, a}^{\mathsf{\,T}} = \mathbf{I}$) and take the matrix product for each term within the MSE loss in eq. (\ref{eq:direct_minilm}), we obtain the following loss function:
\begin{equation}
\sum_{\substack{\alpha \, \in \\ \{ Q,K,V \}}} \sum_{a = 1}^{A_r} \mathrm{MSE} \Bigl( \mathbf{A}_{\alpha, i, a}^S \mathbf{A}_{\alpha, i, a}^{S\mathsf{\,T}}, \mathbf{A}_{\alpha, j, a}^T \mathbf{A}_{\alpha, i, a}^{T\mathsf{\,T}} \Bigl)
\label{eq:mhsa_transfer}
\end{equation}
\noindent
This loss closely resembles MiniLMv2 from eq. (\ref{eq:mha_transfer}) with a minor difference of using MSE loss instead of CE loss with softmax.
Therefore, DirectMiniLM with certain constraints naturally corresponds to MiniLMv2.
The major difference is in whether $\mathbf{A}_{\alpha, i, a}^T$ is transferred directly (with linear mappings) or indirectly (with relation matrices): by comparing these two approaches, we can precisely quantify the advantage of each optimization technique.

\begin{table*}[t!]
\centering
\begin{adjustbox}{max width=0.95\textwidth}
\begin{tabular}{cccccccc}
\hline
 \multirow{2}{*}{\textbf{Model}}& \multirow{2}{*}{\textbf{Architecture}} & \textbf{Monolingual} &  \textbf{Multilingual}& \multicolumn{2}{c}{\textbf{Monolingual Latency}} & \multicolumn{2}{c}{\textbf{Multilingual Latency}} \\
 &  & \textbf{Params} &  \textbf{Params} &\textbf{GPU} & \textbf{CPU} &\textbf{GPU} & \textbf{CPU}   \\
\hline
6L-DistilBERT  & 6, 12, 768, 3072 &66 & 234 &5.98 (0.03) & 33.28 (0.09)& 6.01 (0.06)& 34.02(0.06)\\
6L   & 6, 12, 384, 1536 & 23 & 106 & 5.69 (0.02)& 11.98 (0.07)& 5.99 (0.07)&12.52 (0.06)\\
4L  & 4, 12, 576, 768 & 27 & 153 &3.66 (0.01) &\phantom{0}9.53 (0.04) &3.98 (0.02) &\phantom{0}9.66 (0.05)\\
3L  & 3, 12, 384, 1024 & 16 & 100 & 3.02 (0.01)&\phantom{0}5.41 (0.08) &3.25 (0.01) &\phantom{0}6.01 (0.06) \\
\hline
Teacher &12, 12, 768, 3072 &110 &277&8.69 (0.08) & 64.91 (0.61)&9.47 (0.01)& 66.31 (0.57)\\
\hline
\end{tabular}
\end{adjustbox}
\caption{\label{tab:architecture}
Model Architectures displayed as [$L$, $A_h$, $d_h$, $d_f$].
All parameters are in millions, with the difference in the  monolingual and multilingual parameters due to the vocabulary sizes (30K for monolingual and 252K for multilingual). All latencies are in milliseconds, measured over 5 runs, with standard deviation in parenthesis. }
\end{table*}

\section{Experimental Setup}
\label{sec:experiments}

We explore the task-agnostic knowledge distillation methods under two settings:\footnote{Note that we limit our study to encoder-only models and leave the distillation of decoder-only \citep{radford2019language} or encoder-decoder \citep{lewis-etal-2020-bart} models as future work.}
\begin{enumerate}[topsep=0pt, itemsep=0pt, leftmargin=.2in, parsep=0pt]
    \item Monolingual Distillation: We train English students using the open-source BERT \cite{devlin2019bert} as the teacher. These models are distilled on the same corpus used for pretraining BERT, i.e., English Wikipedia \cite{devlin2019bert} and BookCorpus \cite{BookCorpus}. 
    \item Multilingual Distillation: We train multilingual students using our in-house XLM-RoBERTa \cite{conneau2020unsupervised} as the teacher, and distill on the CC100 dataset \cite{conneau2020unsupervised}, which consists of data in more than 100 languages. We only use a small subset of the corpus to conduct our experiments within a reasonable computation budget while maintaining the language-wise distribution.
\end{enumerate}
\noindent
In both settings, we use the Base (12 layer) architecture for the teacher, as shown in Table \ref{tab:architecture}.
For more details on each distillation setup (e.g. hyperparameters), please refer to Appendix \ref{app:distil}.

\paragraph{Student Models}
To conduct a strong comparison of the representative knowledge distillation methods, we train 4 students of varying architectures and latency/parameter budgets. 
A summary of the student architectures, with their parameters and latency of inference, are shown in Table \ref{tab:architecture}.

Our largest student is a 6 layer model that follows the same architecture as DistilBERT \cite{sanh2019distilbert}. We also use the 6 layer model used in \citet{mukherjee2021xtremedistiltransformers}, which has a smaller hidden size than the teacher. Our smaller 4 and 3 layer students were obtained as recommendations from a Neural Architecture Search process \citep{trivedi2023neural} to find good student architectures for distillation from the XLM-RoBERTa teacher, conditioned to minimize the latency on CPU. Please refer to \cref{app:NAS} for more details.

\paragraph{Layer Mapping Strategies}
The layer mapping strategy $\phi$ is a parameter that needs to be considered for both HS and MHA transfer.
For HS transfer, we explore the following three settings:

\begin{enumerate}[topsep=0pt, itemsep=0pt, leftmargin=.2in, parsep=0pt]
    \item Single Mapping: We only distil the last (${L^T}^{\mathrm{th}}$) teacher layer into the last student layer, which has been shown to be a simple yet competitive baseline \citep{ko2023revisiting}.
    \item 1-to-1 Mapping: Prior work shows that mapping not only the last layer but also the intermediate layers improves distillation \citep{sun-etal-2019-patient}. In 1-to-1 mapping, we distil one teacher layer into each student layer by choosing:
    \begin{itemize}[topsep=0pt, itemsep=0pt, leftmargin=.2in, parsep=0pt]
        \item \emph{Last} $L^S$ teacher layers, i.e. $\phi (i) = \{ L^T - L^S + i \}$ \,($i \in [1,L^S]$). Empirically, last teacher layers capture more high-level (e.g. semantic) knowledge in their representations \citep{tenney2019bert, jawahar-etal-2019-bert}.
        \item A \emph{Uniform} selection of teacher layers which chooses every $k^{\mathrm{th}}$ teacher layer, i.e. $\phi (i) = \{ k i \}$, where $k = \lceil L^T/L^S \rceil$.\footnote{This strategy is used in DistilBERT \cite{sanh2019distilbert} and also known as the "skip" strategy \cite{sun-etal-2019-patient}.} This method can also transfer the lower teacher layers, which empirically captures local (e.g. syntactic) knowledge \citep{tenney2019bert}.
    \end{itemize}
    \item 1-to-N Mapping: Some works even show that mapping each student layer to multiple teacher layers can avoid the loss of information and facilitate student learning \citep{wu-etal-2020-skip,passban2021alp}.
    For 1-to-N Mapping, we explore the following choices of teacher layers:
    \begin{itemize}[topsep=0pt, itemsep=0pt, leftmargin=.2in, parsep=0pt]
        \item A uniform selection of $k$ consecutive layers (\emph{Uniform-Cons.}), i.e. $\phi (i) = [k(i-1), ki]$, where $k = \lceil L^T/L^S \rceil $. This avoids the loss of information since all teacher layers are mapped to at least one student layer.
        \item Combining the \emph{Uniform} and \emph{Last} strategies from the 1-to-1 mapping (\emph{Uniform+Last}). This selects 2 teacher layers per student layer based on each 1-to-1 strategy, expecting to take the best out of both approaches.
    \end{itemize}
\end{enumerate}

\begin{table}[t!]
\centering
\begin{adjustbox}{max width=0.47\textwidth}
\begin{tabular}{cl}
\hline
 \textbf{Distillation Method}  & \textbf{Layer Mapping Strategies} \\
\hline
 &  Single: ${L^T}^{\mathrm{th}}$ \\
HS Transfer  &  1-to-1: Last, Uniform \\
& 1-to-N: Uniform-Cons., Uniform+Last \\
\hline
MHA Transfer & Single: ${L^T}^{\mathrm{th}}$, $(L^T\!-\!1)^{\mathrm{th}}$, $(L^T\!-\!2)^{\mathrm{th}}$ \\
\hline
\end{tabular}
\end{adjustbox}
\caption{\label{tab:distil-options}
Layer mapping strategies explored in each distillation method. The same strategies are explored for MiniLMv2 and DirectMiniLM in MHA Transfer.} 
\end{table}

\begin{table*}[ht!]
    \centering
    \begin{adjustbox}{max width=0.99\textwidth}
    \begin{tabular}{c|c|cccc|cccc|cccc}
    \hline
     \multirow{3}{*}{\minitab[c]{\textbf{Distillation}\\ \textbf{Method}}} &\textbf{Layer}& \multicolumn{4}{c|}{\textbf{Avg. GLUE (Monolingual)}} & \multicolumn{4}{c|}{\textbf{Avg. GLUE (Multilingual)}} & \multicolumn{4}{c}{\textbf{Avg. XNLI (Multilingual)}} \\
      &\textbf{Mapping} & 6L- & \multirow{2}{*}{6L} & \multirow{2}{*}{4L} & \multirow{2}{*}{3L} & 6L- & \multirow{2}{*}{6L} & \multirow{2}{*}{4L} & \multirow{2}{*}{3L} & 6L- & \multirow{2}{*}{6L} & \multirow{2}{*}{4L} & \multirow{2}{*}{3L} \\
     & \textbf{Strategy}& DistilBERT & & & & DistilBERT & & & & DistilBERT & &  & \\
    \hline
    \hline
&${L^T}^{\mathrm{th}}$ &	\underline{84.1}&	79.4&	80.2	&\underline{\textbf{78.9}}& 80.8 &	77.1 &	78.0 &	74.7 & 56.2 &	55.1 &	51.6 &	50.6 \\
& Last &	83.2 &	80.4 &	79.3 &	77.7 & 81.7&	77.0 &	78.3 &	72.6 & 63.1 &	61.0 &	60.3 &	54.4 \\
HS Transfer& Uniform&	82.9 &	\underline{80.6}&	79.6 &	76.6 & 81.6 &78.2 &	78.3 &	73.5 & 59.9 &	59.9 & 59.7 &	\underline{59.9}\\
& Uniform-Cons. &83.9 &	\underline{80.6}&	\underline{80.6}&	77.7 &82.4 &	\underline{78.8}&	78.0 &	\underline{\textbf{75.9} }& 65.5 & 62.2 & 60.4 &	58.6 \\
& Uniform+Last&	\underline{84.1}&	80.4 &	80.4 &	77.7 &\underline{83.1}&	78.7 &	\underline{79.2}&	75.0 & \underline{67.0}&	\underline{62.7}&	\underline{62.5}& 57.9 \\
\hline
&${L^T}^{\mathrm{th}}$&	 \underline{84.1}&	78.1&	79.4 &	76.6 &	78.5 &	75.1 &	75.2 &	67.9 & 50.5 &	48.2 &	51.6 &	43.8 \\
    OD Transfer&Last & 83.1 &	80.4 &	79.3 &	76.4 	&80.7  &	76.9 &	76.1 	&69.8  & 62.6 	&57.0 &	54.1 	&42.7 \\
    (init. from&Uniform& 83.4 	&79.8&	79.8 &	\underline{77.1}	&79.9 &	78.0 	&\underline{77.9}	&65.4 & 60.4 &	54.1 	&52.0 &	42.8 \\
    HS Transfer)&Uniform-Cons.& 83.7 	&80.3 &	79.5 	&76.7 &	81.7&	\underline{78.7}&	76.4 	&70.1 & 63.1 &	\underline{61.0}	&56.5 &	48.2 \\
    &Uniform+Last& \underline{84.1}&	\underline{80.5} &	\underline{79.9}&	\underline{77.1}&	\underline{82.1}&	78.4 &	76.4 	&\underline{72.3}& \underline{66.0} &	60.9 &	\underline{60.0}&	\underline{48.6}\\
\hline
    \multirow{3}{*}{MiniLMv2} & ${L^T}^{\mathrm{th}}$ & 84.2 &	81.9&	79.9&	77.6 &	82.3 &	80.1 &	79.3 &	74.4 & 67.0 &	66.7 	&63.1 	&59.3 	\\
    &$(L^T\!-\!1)^{\mathrm{th}}$& 84.2 &	\underline{\textbf{82.5}}&	80.0 &	78.2 &	\underline{83.1}&	\underline{81.0} &	\underline{80.2} &	\underline{75.8}& \underline{\textbf{69.1} }	& \underline{\textbf{67.5} }&	\underline{\textbf{65.6}}&	\underline{\textbf{62.0}}	\\
    &$(L^T\!-\!2)^{\mathrm{th}}$ & \underline{\textbf{84.4}}&	82.2 &	\underline{\textbf{80.7}}&	\underline{78.3}&	82.9 &	80.5	&78.3 &	73.4 & 67.5 &	66.9 &	63.5 &	61.5 	\\
    \hline
    \multirow{3}{*}{DirectMiniLM} & ${L^T}^{\mathrm{th}}$ & 84.0 & 81.3 & 79.7 & 78.2 & 83.2 & 80.8 & 79.0 & 75.1 & 66.3 & \underline{66.1} & 64.7 & 60.7 \\
    & $(L^T\!-\!1)^{\mathrm{th}}$ & \underline{\textbf{84.4}} & \underline{81.7} & 79.6 & 78.0 & 81.9 & \underline{\textbf{81.1}} & \underline{\textbf{80.3}} & 73.8 & \underline{66.9} & 65.9 & 64.8 & \underline{61.0} \\
    & $(L^T\!-\!2)^{\mathrm{th}}$ & 84.3 & \underline{81.7} & \underline{80.4} & \underline{78.3} & \underline{\textbf{83.4}} & 80.9 & 79.7 & \underline{75.6} & 66.3 & 64.8 & \underline{65.4} & 60.5 \\
\hline
\hline
    \multicolumn{2}{c|}{Teacher} & \multicolumn{4}{c|}{85.5 } & \multicolumn{4}{c|}{84.8 }&  \multicolumn{4}{c}{70.9 } \\
\hline
    \end{tabular}
    \end{adjustbox}
    \caption{\label{tab:all_results}
    Performance of the representative distillation methods evaluated on avg. GLUE and XNLI. Results based on the best layer mapping strategy for each method is underlined, and the best overall result is shown in bold.}
    \end{table*}

For MHA transfer, we always take the single mapping strategy and distill a single teacher layer into the last student layer, following \citet{wang-etal-2021-minilmv2}. Specifically, we experiment with the last three teacher layers as a choice for distillation for both MiniLMv2 and DirectMiniLM. Table \ref{tab:distil-options} summarizes our layer selection options.

While OD transfer can be conducted from scratch, we found this converges slowly and does not perform competitively.\footnote{Our 6L monolingual student takes 49 hours on 30 V100 GPUs to reach acceptable performance, while the same model achieves better scores in only 10.5 hours when initialized from the HS transferred checkpoint.}
Therefore, we take the style of \textit{multi-stage} distillation \citep{mukherjee2021xtremedistiltransformers} and conduct OD transfer after HS transfer, using the distilled checkpoint from HS transfer.
This approach converges much faster with better final performance, hence we take this approach as the representative OD transfer method.

\section{Evaluation and Results}
\label{sec:evaluation}

For both our monolingual and multilingual models, we measure performance on the English GLUE Benchmark \cite{wang2019glue} and report the average score of all tasks (without CoLA\footnote{Distilled models often perform poorly on CoLA: We hypothesize this is because CoLA is the only \emph{syntactic} task in the benchmark as opposed to the other \emph{semantic} tasks \cite{xu2022autodistil}. We include the results of CoLA in \cref{app:extended-eval}.}).
For multilingual models, we provide evaluations on the XNLI dataset \cite{conneau2018xnli}, a set of inference tasks which evaluates the model's performance on 15 languages after being finetuned on only English training data. We report the average score of all languages for XNLI.

\cref{tab:all_results} summarizes the performance of each distillation method on 4 student architectures.
For detailed evaluations of each method based on the best configuration, please refer to \cref{app:extended-eval}.
We also provide a comparison against DistilBERT \citep{sanh2019distilbert}, a representative \textit{architecture-constrained} method, in \cref{app:distilbert}.

\paragraph{HS Transfer} From \cref{tab:all_results}, we can verify that the performance of HS transfer varies with different layer mapping strategies, and no strategy dominates the others in all settings.
In the monolingual setting, we found that the single mapping strategy performs competitively, which is in line with the findings of \citet{ko2023revisiting}.
However, in the multilingual setting, more sophisticated 1-to-N strategies generally show superiority over the simpler baselines.
This indicates that more supervision from the teacher can be helpful (and at worst harmless), hence we advocate for the adoption 1-to-N strategies, esp. in the challenging multilingual distillation.

\paragraph{OD Transfer} As mentioned in \cref{sec:experiments}, we initialize the model from the HS transferred checkpoints with each layer mapping strategy.
Interestingly, we see a slight \emph{degradation} in performance on downstream tasks compared to only HS transfer, with a significant loss observed for smaller students.
This indicates that learning effective representations from the output distribution signals is difficult, especially for students with lower capacity.
Moreover,  given how computationally expensive OD transfer can be, HS transfer is a cheaper and more effective alternative for knowledge transfer. 

\paragraph{MHA Transfer} For both MiniLMv2 and DirectMiniLM, we found distilling the upper-middle teacher layer, i.e. $(L^T\!-\!1)^{\mathrm{th}}$ or $(L^T\!-\!2)^{\mathrm{th}}$ strategy, led to the best performance, in line with the original findings of \citet{wang-etal-2021-minilmv2}.
Importantly, we found that both MHA transfer methods generally outperform HS transfer, which points to the benefit of transferring the Q/K/V knowledge over the hidden state knowledge. This is consistent with the latest comparative study by \citet{wang2023distill}, although they only evaluate on the 6L-DistilBERT architecture in the monolingual setting.

We also note that MiniLMv2 and DirectMiniLM perform equivalently, with the notable exception on XNLI.
We attribute this to two factors: 
\begin{enumerate}[topsep=0pt, itemsep=0pt, leftmargin=.2in, parsep=0pt]
\item MiniLMv2 transfers relational representations conditioned on the whole input, while DirectMiniLM transfers absolute position-wise representations.
The former may be more semantically informative, as the contextual representations often exhibit rich relational structures \citep{park-etal-2021-distilling,liu-etal-2022-multi-granularity}.
\item DirectMiniLM requires learning the linear transformation weight $\mathbf{W}_{\alpha, a}$, while MiniLMv2 does not incur any additional parameters.
\end{enumerate}
From these observations, we generally expect MiniLMv2 to be the best distillation method and have adopted it in our latency-critical applications.\footnote{Specifically, the 4L monolingual and multilingual students with 7x speedup on CPU have been deployed for various NLP applications, such as entity extraction, document classification and relation detection, while maintaining 93\% of the teacher's performance on average \cite{trivedi2023neural}.}
However, DirectMiniLM performs comparably and provides meaningful insights on the benefit of each optimization technique, which can be useful for debugging and analyzing MiniLMv2.
Therefore, we recommend its comparison for both reseachers and practitioners in future studies.

\section{Conclusion}
\label{sec:conclusion}

This study critically analyzes the representative methods for task-agnostic distillation of language models.
Specifically, we compare Output Distribution (OD), Hidden State (HS), and Multi-Head Attention (MHA) transfer for different student architectures, language settings, and layer mapping strategies.
Through our extensive experiments, we show that MHA transfer based on MiniLMv2 is the best option across many settings, followed by HS transfer with sophisticated 1-to-N mapping strategies. Meanwhile, we did not find OD transfer to be an effective alternative.
Finally, we propose DirectMiniLM to demistify the precise advantage of the indirect (i.e. relation matrix based) optimization technique proposed in MiniLMv2.
Overall, we hope this study will be a useful guide for both researchers and practitioners working in this area.

\bibliography{anthology,custom}
\bibliographystyle{acl_natbib}

\appendix

\section{Related Work}
\label{sec:related_work}

MobileBERT \citep{sun-etal-2020-mobilebert} is an effective technique to compress BERT into a specially designed student with a bottleneck architecture.
In BERT-of-Theseus \citep{xu-etal-2020-bert}, the modules of the teacher are progressively replaced with smaller ones to improve efficiency.
However, these approaches constrain the architecture of the students.
In contrast, we focus on the \textit{architecture-agnostic} distillation methods for better flexibility.

Improvements on distillation objectives are also made, e.g. transferring the relational, structural or holistic representations of the language models may provide more useful signals for students \cite{park-etal-2021-distilling,liu-etal-2022-multi-granularity,tan-etal-2023-multilingual}.
When the transfer set is limited, various methods of data augmentation \citep{liang2021mixkd,Zhang_Naresh_He_2022,liu-etal-2022-rethinking-task} can be applied successfully.
To alleviate the \textit{capacity gap} between the teacher and student, previous works proposed scheduled annealing in OD transfer \citep{jafari-etal-2021-annealing}, multi-stage distillation with intermediate-sized teacher assistants \citep{mirzadeh2020improved,son2021densely}, and meta-learning to optimize the teacher for student distillation \citep{zhou-etal-2022-bert,ma-etal-2022-knowledge}.
We leave the exploration of such advanced techniques as future work.

Layer mapping strategies for HS transfer have also been studied extensively.
\citet{jiao2021improving} proposed an evolutionary search process to obtain the optimal layer mapping for specific downstream tasks.
\citet{li-etal-2020-bert} applied Earth Mover's Distance to prioritize mappings with smaller cost (i.e. distillation loss).
The attention mechanism can also be applied to map student layers to \textit{similar} teacher layers, where the similarity is computed based on the cosine similarity \citep{passban2021alp} or the predictions of internal classifiers
\citep{wu-etal-2021-universal}.
Finally, random mapping has been shown to work surprisingly well, potentially working as a regularizer to prevent overfitting \citep{haidar-etal-2022-rail}.
In this study, we focus instead on the carefully designed and easily applicable heuristic strategies.

Finally, there are different approaches to reducing the inference costs of large language models, such as quantization \citep{zafrir2019q8bert,shen2020q,kim2021bert,bai-etal-2021-binarybert}, pruning \citep{Fan2020Reducing,lagunas-etal-2021-block,xia-etal-2022-structured}, early exit mechanisms \citep{liu-etal-2020-fastbert,xin-etal-2021-berxit,liao-etal-2021-global,wang-etal-2022-skipbert}, and matrix decomposition \citep{ben-noach-goldberg-2020-compressing,mao-etal-2020-ladabert,NEURIPS2021_chen,tahaei-etal-2022-kroneckerbert}.
Many of these approaches are complementary to our distillation methods and can be combined for further efficiency.

\section{Distillation Setup}
\label{app:distil}

We train our monolingual students on the entire Wikipedia and BookCorpus using the AdamW Optimizer \cite{loshchilov2019adamw} with $\beta_1 = 0.9, \beta_2=0.98$. For HS and MHA transfer, students are trained for $7$ epoch with a peak learning rate (LR) of $5e-4$. For OD transfer, we train for $3$ epochs with a peak LR of $3e-4$ after HS transfer. We use a linear LR warmup over the first $5\%$ of the training steps and then a linear decay.
 We use a batch size of $32$ with the maximum sequence length set to 256 and train on 30 V100 GPUs.
 
For multilingual distillation, we use a small subset of CC-100 containing 7M sentences, which we found to be sufficient for developing competitive students.
We generally use the same setup as monolingual distillation, except we use the peak LR of $8e-4$ for MHA transfer. Multilingual students are trained on 2 A100-80GB GPUs.

Finally, the method-specific hyperparameters (\cref{sec:distillation_methods}) are as follows.
For OD transfer, we set the output temperature $\mathcal{T}$ to the default value of $1$.
For MiniLMv2, we use $A_r > A_h$ to transfer more fine-grained knowledge in the Q/K/V mappings: specifically, we set $A_r=48$, which is also used in \citet{wang-etal-2021-minilmv2}.
For DirectMiniLM, we found using $A_r=A_h$ without the orthogonal constraints on $\mathbf{W}_{\alpha, a}$ led to the best performance and used this setting throughout our experiments.

\section{Finding Smaller Student Models}
\label{app:NAS}
Our smallest students, a 4 layer and a 3 layer model, were obtained as recommendations from a Neural Architecture Search process to find good student architectures for task-agnostic distillation from an XLM-RoBERTa teacher, conditioned to minimize the latency of inference on a CPU. Specifically, we follow the KD-NAS method of \citet{trivedi2023neural} and modify the reward to reduce the distillation loss $\mathcal{L}_{\mathrm{HS}}$ defined in \cref{eq:hs_transfer}, along with the CPU latency of the student ($lat(S)$) normalized by the teacher's latency ($lat(T)$):
\begin{equation}
reward(S) = (1-\mathcal{L}_{\mathrm{HS}})  * \left(\frac{lat(S)}{0.6 * lat(T) }\right)^{-0.06}
\end{equation}
\label{eq:reward}
Please refer to their original paper for more details.

\section{Evaluation Results for Best Models}
\label{app:extended-eval}

We include detailed results of each distillation method for the best configuration (i.e. layer mapping strategy).
Specifically, we show the results of each GLUE task for monolingual and multilingual distillation in
\cref{tab:extended-glue-mono} and \ref{tab:extended-glue-multi}. We show language-wise performance on XNLI in \cref{tab:extended-xnli}.
All downstream tasks are evaluated on 3 random seeds.

For the sake of efficient evaluation, we did not conduct expensive grid search for finetuning hyperparameters. After some manual tuning, we used the same LR of $2e-5$ and batch size of 32 for finetuning all models on all tasks. We used 3 epochs of finetuning for GLUE tasks (except CoLA, where we used 6 and 10 epochs for monolingual and multilingual models) and 5 epochs for XNLI.

\section{Architecture Constrained Distillation: DistilBERT}
\label{app:distilbert}

DistilBERT \citep{sanh2019distilbert} is one of the earliest and most widely used baseline.
This method comprises (1) layer initialization from the teacher layers, (2) HS transfer based on cosine similarity loss, and (3) OD transfer.
The first two techniques restrict the architecture of each student layer to be identical to the teacher model, which limits our analysis to the 6L-DistilBERT student architecture.

\begin{table}[ht!]
    \centering
    \begin{adjustbox}{max width=0.48\textwidth}
    \begin{tabular}{ccc}
    \hline
    & 6L-DsitilBERT & Teacher \\
    \hline
    Avg. GLUE (Monolingual) & 82.9 (0.5) &85.5 (0.6) \\
    Avg. GLUE (Multilingual) & 79.7 (0.5) & 84.8 (0.3) \\
    Avg. XNLI (Multilingual) & 61.8 (0.5) & 70.9 (0.8) \\
    \hline
    \end{tabular}
    \end{adjustbox}
    \caption{\label{tab:DistilBERT}
    DistilBERT Performance. Average GLUE scores reported for all tasks w/o CoLA. Average XNLI scores reported for all languages. Average taken over 3 random seeds with standard deviation in parenthesis.}
    \end{table}

As shown in the results of \cref{tab:DistilBERT}, the performance of DistilBERT is generally not competitive with our distillation methods from \cref{tab:all_results}, especially in the multilingual setting.

\begin{table*}[ht!]
\centering
\begin{adjustbox}{max width=\textwidth}
\begin{tabular}{c|c|c|cccccccc|cc}
\hline
\multirow{2}{*}{\textbf{Model}} & \textbf{Distillation} & \textbf{Best} & \multicolumn{8}{c|}{\textbf{GLUE Performance}} & \multirow{2}{*}{\textbf{Avg.}} & \textbf{Avg.} \\
 & \textbf{Method} & \textbf{Strategy} & \textbf{MNLI} & \textbf{QQP} & \textbf{QNLI} & \textbf{SST-2} & \textbf{CoLA} & \textbf{STS-B} & \textbf{MRPC} & \textbf{RTE} &  & \textbf{(-CoLA)} \\
\hline
\hline
\multirow{4}{*}{6L-DistilBERT} & HS Transfer & Uniform+Last & 82.6 & 86.2 & 88.7 & 90.8 & 45.9 & 85.9 & 89.7 & 65.1 & 79.4 (0.5) & 84.1 (0.4) \\
 & OD Transfer & Uniform+Last & 82.7 & 86.5 & 88.3 & 91.3 & 50.8 & 85.5 & 89.7 & 64.4 & 79.9 (0.3) & 84.1 (0.2) \\
 & MiniLMv2 & $(L^T\!-\!2)^{\mathrm{th}}$ & 83.0 & 86.6 & 90.1 & 91.6 & 53.1 & 86.7 & 89.0 & 64.2 & \textbf{80.5 (0.4)} & \textbf{84.4 (0.3)} \\
 & DirectMiniLM & $(L^T\!-\!1)^{\mathrm{th}}$ & 82.9 & 86.6 & 90.0 & 91.4 & 52.7 & 86.4 & 89.0 & 64.9 & \textbf{80.5 (0.5)} & \textbf{84.4 (0.4)} \\
 \hline
 \multirow{4}{*}{6L} & HS Transfer & Uniform-Cons. & 78.3 & 85.0 & 85.9 & 90.9 & 31.2 & 83.2 & 84.4 & 56.3 & 74.4 (0.4) & 80.6 (0.3) \\ 
  & OD Transfer & Uniform+Last & 79.1 & 84.6 & 86.3 & 89.7 & 38.6 & 82.3 & 83.7 & 57.9 & 75.3 (0.6) & 80.5 (0.3) \\
  & MiniLMv2 & $(L^T\!-\!1)^{\mathrm{th}}$ & 80.8 & 84.9 & 88.0 & 90.3 & 36.2 & 84.5 & 86.2 & 62.5 & \textbf{76.7 (0.1)} & \textbf{82.5 (0.1)} \\
  & DirectMiniLM & $(L^T\!-\!1)^{\mathrm{th}}$ & 80.0 & 85.1 & 87.2 & 90.9 & 36.1 & 83.3 & 85.9 & 59.7 & 76.0 (0.2) & 81.7 (0.2) \\
 \hline
\multirow{4}{*}{4L} & HS Transfer & Uniform-Cons. & 77.3 & 84.9 & 85.7 & 90.0 & 26.9 & 83.4 & 83.0 & 60.1 & 73.9 (0.4) & 80.6 (0.3) \\
  & OD Transfer & Uniform+Last & 78.2 & 84.6 & 85.1 & 90.1 & 32.2 & 83.3 & 83.2 & 55.1 & 74.0 (0.2) & 79.9 (0.4) \\
  & MiniLMv2 & $(L^T\!-\!2)^{\mathrm{th}}$ & 78.8 & 83.8 & 86.0 & 90.8 & 30.9 & 83.0 & 84.3 & 58.2 & \textbf{74.5 (0.2)} & \textbf{80.7 (0.3)} \\
  & DirectMiniLM & $(L^T\!-\!2)^{\mathrm{th}}$ & 79.0 & 84.2 & 85.7 & 90.0 & 29.7 & 82.5 & 84.9 & 56.6 & 74.1 (0.4) & 80.4 (0.4) \\
 \hline
 \multirow{4}{*}{3L} & HS Transfer & ${L^T}^{\mathrm{th}}$ & 74.3 & 82.8 & 84.0 & 89.4 & 20.0 & 80.8 & 83.4 & 57.5 & \textbf{71.5 (0.1)} & \textbf{78.9 (0.3)} \\
  & OD Transfer & Uniform+Last & 73.8 & 81.9 & 83.4 & 86.6 & 15.1 & 78.8 & 82.7 & 52.8 & 69.4 (0.3) & 77.1 (0.4) \\
  & MiniLMv2 & $(L^T\!-\!2)^{\mathrm{th}}$ & 75.1 & 81.9 & 84.8 & 87.3 & 13.3 & 81.6 & 82.0 & 55.1 & 70.1 (0.4) & 78.3 (0.2) \\
  & DirectMiniLM & $(L^T\!-\!2)^{\mathrm{th}}$ & 75.7 & 82.2 & 84.0 & 88.5 & 16.8 & 81.0 & 83.3 & 53.5 & 70.6 (0.2) & 78.3 (0.3) \\
  \hline
  \hline
\multicolumn{3}{c|}{Teacher} & 84.4 & 88.0 & 91.5 & 92.9 & 57.4 & 88.0 & 89.0 & 64.8 & 82.0 (0.6) & 85.5 (0.6)\\
  \hline
\end{tabular}
\end{adjustbox}
\caption{\label{tab:extended-glue-mono}
    Monolingual Student GLUE Performance for all tasks. Each row shows performance based on the best layer mapping strategy. Each score reported as an average over 3 random seeds (standard deviation in parenthesis). }
\end{table*}

\begin{table*}[ht!]
\centering
\begin{adjustbox}{max width=\textwidth}
\begin{tabular}{c|c|c|cccccccc|cc}
\hline
\multirow{2}{*}{\textbf{Model}} & \textbf{Distillation} & \textbf{Best} & \multicolumn{8}{c|}{\textbf{GLUE Performance}} & \multirow{2}{*}{\textbf{Avg.}} & \textbf{Avg.} \\
 & \textbf{Method} & \textbf{Strategy} & \textbf{MNLI} & \textbf{QQP} & \textbf{QNLI} & \textbf{SST-2} & \textbf{CoLA} & \textbf{STS-B} & \textbf{MRPC} & \textbf{RTE} &  & \textbf{(-CoLA)} \\
\hline
\hline
\multirow{4}{*}{6L-DistilBERT} & HS Transfer & Uniform+Last & 80.8 & 86.8 & 87.9 & 90.2 & 32.3 & 84.7 & 88.5 & 62.6 & 76.7 (0.6) & 83.1 (0.3) \\
 & OD Transfer & Uniform+Last & 80.1 & 86.4 & 86.2 & 89.8 & 33.1 & 84.1 & 87.5 & 60.5 & 76.0 (1.0) & 82.1 (0.5) \\
 & MiniLMv2 & $(L^T\!-\!1)^{\mathrm{th}}$ & 81.3 & 85.8 & 88.8 &	89.6 & 40.2 & 85.9 & 89.3 & 61.0 & \textbf{77.7 (0.5)} & 83.1 (0.3) \\
 & DirectMiniLM & $(L^T\!-\!2)^{\mathrm{th}}$ &	81.0&	86.4&	89.2&	89.8&	37.8&	85.9&	90.1&	61.7&	\textbf{77.7 (0.7)}&	\textbf{83.4 (0.6)} \\
 \hline
 \multirow{4}{*}{6L} & HS Transfer & Uniform-Cons. & 75.0 & 82.8 & 83.0 & 86.7 & 16.9 & 80.8 & 84.6 & 58.5 & 71.1 (0.6) & 78.8 (0.4) \\
 & OD Transfer & Uniform-Cons. & 76.2 & 83.7 & 83.6 & 87.5 & 16.9 & 78.1 & 85.0 & 55.9 & 71.1 (0.6) & 78.7 (0.5) \\
  & MiniLMv2 & $(L^T\!-\!1)^{\mathrm{th}}$ &	78.3 &	83.7&	86.9& 89.1 &	29.2&	83.6&	85.1&	60.3 &	\textbf{74.5 (0.5)} &	81.0 (0.4) \\
  & DirectMiniLM & $(L^T\!-\!1)^{\mathrm{th}}$ &	78.3 &	84.3&	86.1&	89.4&	25.5&	84.5&	86.9 &	58.0 &	74.1 (0.6)&	\textbf{81.1 (0.5)} \\
 \hline
\multirow{4}{*}{4L} & HS Transfer & Uniform+Last & 75.6 & 83.7 & 83.8 & 87.8 & 18.3 & 81.2 & 83.3 & 59.0 & 71.6 (0.7) & 79.2 (0.5) \\
  & OD Transfer & Uniform & 73.4 & 83.8 & 81.2 & 85.2 & 17.0 & 80.0 & 82.8 & 58.6 & 70.3 (0.7) & 77.9 (0.7) \\
  & MiniLMv2 & $(L^T\!-\!1)^{\mathrm{th}}$ & 76.8 & 83.4 & 85.2 & 87.6 & 17.1 & 83.9 & 86.0 & 58.1 & \textbf{72.3 (0.7)} & 80.2 (0.5) \\
  & DirectMiniLM & $(L^T\!-\!1)^{\mathrm{th}}$ & 77.0 & 83.6 & 85.2 & 88.5 & 19.2 & 83.5 & 85.2 & 59.1 & 72.7 (0.6) & \textbf{80.3 (0.4)} \\
 \hline
 \multirow{4}{*}{3L} & HS Transfer & Uniform-Cons. & 71.0 & 80.7 & 82.1 & 84.6 & 11.0 & 75.8 & 82.2 & 54.9 & 67.8 (0.4) & \textbf{75.9 (0.4)} \\
  & OD Transfer & Uniform+Last & 68.1 & 79.4 & 79.7 & 81.9 & 2.6 & 61.5 & 81.2 & 54.6 & 63.6 (0.5) & 72.3 (0.6) \\
  & MiniLMv2 & $(L^T\!-\!1)^{\mathrm{th}}$ & 72.7 & 80.6 & 83.2 & 84.6 & 9.7 & 70.6 & 81.7 & 57.4 & 67.6 (0.6) & 75.8 (0.5) \\
  & DirectMiniLM & $(L^T\!-\!2)^{\mathrm{th}}$ & 72.2 & 81.2 & 83.4 & 84.8 & 15.9 & 67.9 & 82.0 & 58.0 & \textbf{68.2 (1.1)} & 75.6 (1.1) \\
  \hline
  \hline
 \multicolumn{3}{c|}{Teacher} & 84.1 & 87.9 & 90.2 & 91.9 & 51.7 & 86.6 & 91.4 & 61.4 & 80.6 (0.3) & 84.8 (0.3) \\
  \hline
\end{tabular}
\end{adjustbox}
\caption{\label{tab:extended-glue-multi}
    Multilingual Student GLUE Performance for all tasks. Each row shows performance based on the best layer mapping strategy. Each score reported as an average over 3 random seeds (standard deviation in parenthesis). }
\end{table*}

\begin{table*}[ht!]
\centering
\begin{adjustbox}{max width=\textwidth}
\begin{tabular}{c|c|c|ccccccccccccccc|c}
\hline
\multirow{2}{*}{\textbf{Model}} & \textbf{Distillation} & \textbf{Best} & \multicolumn{15}{|c|}{\textbf{XNLI Performance}} & \multirow{2}{*}{\textbf{Avg.}} \\
 & \textbf{Method} & \textbf{Strategy} & \textbf{ar} & \textbf{bg} & \textbf{de} & \textbf{el} & \textbf{en} & \textbf{es} & \textbf{fr} & \textbf{hi} & \textbf{ru} & \textbf{sw} & \textbf{th} & \textbf{tr} & \textbf{ur} & \textbf{vi} & \textbf{zh} &  \\
\hline
\hline
\multirow{4}{*}{6L-DistilBERT} & HS Transfer & Uniform+Last & 64.7 & 69.7 & 69.6 & 69.2 & 80.7 & 72.0 & 70.2 & 64.6 & 67.7 & 51.2 & 65.3 & 62.5 & 58.9 & 70.4 & 68.6 & 67.0 (0.4) \\
 & OD Transfer & Uniform+Last & 63.7 & 69.1 & 69.4 & 67.0 & 78.6 & 70.7 & 68.9 & 60.0 & 69.0 & 51.2 & 65.4 & 61.9 & 57.9 & 68.5 & 68.8 & 66.0 (0.6) \\
 & MiniLMv2 & $(L^T\!-\!1)^{\mathrm{th}}$ & 65.5 & 71.6 & 72.1 & 71.5 & 81.4 & 75.0 & 73.5 & 65.3 & 70.6 & 58.1 & 65.1 & 67.1 & 60.9 & 69.7 & 69.3 & \textbf{69.1 (0.5)} \\
 & DirectMiniLM & $(L^T\!-\!1)^{\mathrm{th}}$ & 63.8 & 69.4 & 69.3 & 68.5 & 79.2 & 73.2 & 71.2 & 64.1 & 67.2 & 55.1 & 63.9 & 65.6 & 59.7 & 66.6 & 67.0 & 66.9 (0.4) \\
\hline
\multirow{4}{*}{6L} & HS Transfer & Uniform+Last & 59.7 & 67.2 & 63.4 & 65.6 & 75.9 & 68.7 & 66.8 & 58.3 & 62.4 & 48.9 & 62.7 & 59.1 & 53.4 & 63.2 & 65.1 & 62.7 (0.4) \\
 & OD Transfer & Uniform+Last & 55.7 & 62.6 & 63.7 & 59.2 & 76.5 & 66.9 & 63.7 & 54.1 & 62.0 & 45.7 & 57.9 & 56.3 & 51.0 & 62.8 & 62.2 & 61.0 (0.5) \\
 & MiniLMv2 & $(L^T\!-\!1)^{\mathrm{th}}$ & 65.0 & 69.7 & 70.4 & 68.8 & 80.3 & 73.1 & 71.5 & 62.9 & 69.3 & 53.8 & 65.0 & 65.7 & 59.6 & 69.2 & 68.0 & \textbf{67.5 (0.5)} \\
 & DirectMiniLM & ${L^T}^{\mathrm{th}}$ & 63.2 & 68.8 & 70.1 & 68.1 & 78.4 & 70.5 & 70.0 & 62.2 & 66.6 & 52.4 & 64.6 & 64.0 & 59.1 & 66.2 & 66.9 & 66.1 (0.5) \\
\hline
\multirow{4}{*}{4L} & HS Transfer & Uniform+Last & 56.9 & 64.5 & 66.2 & 66.3 & 77.3 & 68.2 & 63.9 & 57.9 & 63.9 & 49.2 & 61.8 & 59.2 & 54.0 & 64.2 & 64.2 & 62.5 (0.5) \\ 
 & OD Transfer & Uniform+Last & 55.7 & 62.6 & 63.7 & 59.2 & 76.5 & 66.9 & 63.7 & 54.1 & 62.0 & 45.7 & 57.9 & 56.3 & 51.0 & 62.8 & 62.2 & 60.0 (0.5) \\
 & MiniLMv2 & $(L^T\!-\!1)^{\mathrm{th}}$ & 62.9 & 67.5 & 67.8 & 68.2 & 77.8 & 70.7 & 68.2 & 62.4 & 67.0 & 51.0 & 63.6 & 64.7 & 57.7 & 67.2 & 67.4 & \textbf{65.6 (0.8)} \\
 & DirectMiniLM & $(L^T\!-\!2)^{\mathrm{th}}$ & 63.2 & 68.3 & 67.9 & 67.6 & 78.3 & 69.7 & 69.6 & 63.1 & 64.9 & 49.0 & 64.2 & 62.4 & 58.6 & 67.2 & 66.3 & 65.4 (0.7)\\
\hline
\multirow{4}{*}{3L} & HS Transfer & Uniform & 58.3 & 63.4 & 60.5 & 60.6 & 74.1 & 65.6 & 61.6 & 56.6 & 61.4 & 46.7 & 57.3 & 55.9 & 51.8 & 61.1 & 63.1 & 59.9 (0.5) \\
 & OD Transfer & Uniform+Last & 45.6 & 52.3 & 48.7 & 47.8 & 69.9 & 55.0 & 49.4 & 42.9 & 47.3 & 40.9 & 46.3 & 44.4 & 41.6 & 49.7 & 47.8 & 48.6 (0.5) \\
 & MiniLMv2 & $(L^T\!-\!1)^{\mathrm{th}}$ & 60.0 & 64.9 & 63.6 & 64.3 & 74.1 & 66.7 & 64.2 & 58.2 & 61.8 & 49.4 & 59.7 & 60.7 & 55.3 & 64.2 & 62.4 & \textbf{62.0 (0.8)} \\
 & DirectMiniLM & $(L^T\!-\!1)^{\mathrm{th}}$ & 57.4 & 63.0 & 64.1 & 63.3 & 74.3 & 66.1 & 65.1 & 57.2 & 62.1 & 46.7 & 56.7 & 58.1 & 55.2 & 63.6 & 61.8 & 61.0 (0.4) \\
 \hline
 \hline
\multicolumn{3}{c|}{Teacher} & 69.1 & 73.2 & 74.1 & 72.2 & 83.4 & 75.1 & 73.1 & 69 & 71.3 & 57.3 & 69.7 & 67.7 & 64.1 & 70.8 & 73.3 & 70.9 (0.8) \\
  \hline
\end{tabular}
\end{adjustbox}
\caption{\label{tab:extended-xnli}
    Multilingual Student XNLI Performance for 15 languages. Each row shows performance based on the best layer mapping strategy. Each score reported as an average over 3 random seeds (standard deviation in parenthesis). }
\end{table*}

\end{document}